\documentclass{article}

\usepackage{arxiv}

\usepackage[utf8]{inputenc} % allow utf-8 input
\usepackage[T1]{fontenc}    % use 8-bit T1 fonts
\usepackage{hyperref}       % hyperlinks
\usepackage{url}            % simple URL typesetting
\usepackage{booktabs}       % professional-quality tables
\usepackage{amsfonts}       % blackboard math symbols
\usepackage{nicefrac}       % compact symbols for 1/2, etc.
\usepackage{microtype}      % microtypography
\usepackage{lipsum}
\usepackage{graphicx}
\usepackage{threeparttable}

\usepackage{newunicodechar}
\newunicodechar{ﬁ}{fi}
\newunicodechar{ﬀ}{ff}

\title{Deep-Learning-Based Aerial Image Classification for Emergency Response Applications using Unmanned Aerial Vehicles}

\author{
  Christos Kyrkou\thanks{ckyrkou@gmail.com, https://sites.google.com/site/chriskyrkou/
  } \\
  KIOS Research and Innovation Center of Excellence\\
  University of Cyprus\\
  1 Panepistimiou Avenue, Nicosia Cyprus \\
  \texttt{kyrkou.christos@ucy.ac.cy} \\
   \And
 Theocharis Theocharides \\
  Department of Electrical and Computer Engineering\\
   KIOS Research and Innovation Center of Excellence\\
  University of Cyprus\\
  1 Panepistimiou Avenue, Nicosia Cyprus \\
  \texttt{ttheocharides@ucy.ac.cy} \\
}

\begin{document}
\maketitle

\begin{abstract}
Unmanned Aerial Vehicles (UAVs), equipped with camera sensors can facilitate enhanced situational awareness for many emergency response and disaster management applications since they are capable of operating in remote and difficult to access areas. In addition, by utilizing an embedded platform and deep learning UAVs can autonomously monitor a disaster stricken area, analyze the image in real-time and alert in the presence of various calamities such as collapsed buildings, flood, or fire in order to faster mitigate their effects on the environment and on human population. To this end, this paper focuses on the automated aerial scene classification of disaster events from on-board a UAV. Specifically, a dedicated Aerial Image Database for Emergency Response (AIDER) applications is introduced and a comparative analysis of existing approaches is performed. Through this analysis a lightweight convolutional neural network (CNN) architecture is developed, capable of running efficiently on an embedded platform achieving $\sim3\times$ higher performance compared to existing models with minimal memory requirements with less than $2\%$ accuracy drop compared to the state-of-the-art. These preliminary results provide a solid basis for further experimentation towards real-time aerial image classification for emergency response applications using UAVs.
\end{abstract}

% keywords can be removed

\section{Introduction}

Over the past few years Unmanned Aerial Vehicles (UAVs) have gained considerable interest as a remote sensing platform for various practical applications, such as traffic monitoring \cite{Kyrkou:ICCE:2018} and search and rescue \cite{PetridesICUAS2017}. Recent technological advances such as the integration of camera sensors provide the opportunity for new UAV applications such as monitoring and identifying hazards and disasters in emergency situations (e.g., fire spots in forested areas, flooding threat, road collisions, landslide prone areas) \cite{CV_UAV_Survey:Abdulla:2018} by means of analyzing the captured aerial images in real-time.  In addition, due to their small size UAVs offer fast deployment and can thus provide a unique tool to rapidly assess a situation and improve risk assessment mitigation \cite{PetridesSmartCities2017}. However, there is a unique set of constraints that need to be addressed due to the fact that a UAV has to operate in disaster-stricken areas which often have limited connectivity and visibility to the operators. In such cases an autonomous UAV relies heavily on its on-board sensors and microprocessors to carry out a given task without requiring the feed to be send to a central ground station \cite{PetridesICUAS2017}. The challenge in such cases is to enable the efficient visual processing on-board the UAV given that the available hardware may have limitations in terms of computing power and memory.

Deep learning algorithms such as Convolutional Neural Networks (CNNs) have been widely recognized as a prominent approach for many computer vision applications (image/video recognition, detection, and classification) and have shown remarkable results in many applications \cite{SateliteImageClassification:Maggiori:2016,Hohman:2018:VisualAnalyticsDeepLearning:TVCG,Cheng:WhenDLmeetsML:2018:TGRS}. Hence, there are many benefits stemming from using deep learning techniques in emergency response and disaster management applications to retrieve critical information in a timely-fashion and enable better preparation and reaction during time-critical situations and support the decision-making processes \cite{Nguyen2016ApplicationsOO}. Even though CNNs have been increasingly successful at various classification tasks through transfer learning \cite{Razavian:CNNtransfer:CVPRW:2014}, their inference speed on embedded platforms such as those found on-board UAVs is hindered by the high computational cost that they incur and the model size of such networks is prohibitive from a memory perspective for such embedded devices \cite{DATELowPowerImage2018,Shafique.RoadMap.DATE2018}. However, for many applications local embedded processing near the sensor is preferred over the cloud due to privacy and latency concerns, or operation in remote areas where there is limited or even no connectivity.

This work addresses the problem of on-board aerial scene classification for emergency response applications which is to automatically assign a semantic label to characterize the aerial image that the UAV captures \cite{Wang:ASC:TGRS:2017}. These labels correspond to a danger or hazard that has occurred. The specific use-case under consideration is that a UAV will follow a predetermined path as shown in\cite{PetridesICUAS2017}) and will continuously analyze the frames it receives from the camera through its embedded platform and alert for any potential hazards or dangerous situations that it recognizes. The objective of this work is to enhance the real-time perception capabilities in such scenarios through the development of a CNN model that provides the best trade-off between accuracy and performance and can operate on embedded hardware that is on-board the UAV.

The main contributions of this work are summarized as follows:
\begin{itemize}
    \item Construction of a dedicated database for the application of aerial image classification for emergency response and development of a suitable CNN training strategy.
    \item Development of a CNN (referred to as \textit{ERNet}) with low-computational and suitable for low-cost low-power devices.
\end{itemize}

Through the analysis of the different models and techniques as well as evaluation on a real experimental UAV platform we demonstrate the effectiveness of this approach to simultaneously provide near state-of-the-art accuracy ($\sim90\%$) while being $3\times$ faster on an embedded platform.

\section{Background and Related Work}\label{sec:background}

\subsection{Convolutional Neural Network Architectures}\label{subsec:CNNs}

In the last decade, a lot of progress has been made on CNN-based classification systems. Numerous architectures have been proposed by the deep learning community fuelled by the need to perform even better in image classification tasks such as the ImageNet Large Scale Visual Recognition Competition (ILSVRC). Some of the most important architectures are highlighted next, whose components and ideas that will be used to develop an efficient CNN for embedded aerial scene classification with UAVs.

\textbf{VGG16 \cite{VGG.2014}:} The VGG network has become a popular choice when extracting CNN features from images. This particular network contains $16$ CONV/FC layers and appealingly, is characterized by its simplicity. It is comprised only of $3\times 3$ convolutional layers stacked on top of each other with an increasing depth of $2$ with pooling layers in between to reduce the feature map size by a factor of $2$; and with $2$ fully-connected layers at the end, each with $4,096$ neurons. A final dense layer is equal to the number of classes is used for the final classification. A downside of the VGGNet is that it is more expensive to evaluate, and uses a lot of parameters and consequently memory ($\sim140$MB). 

\textbf{ResNet \cite{ResNet.2015}:} This network introduced the idea of residual learning in order to train even deeper CNNs, where the input to a convolution layer is propagated and added to the output of that layer after the operation, thus the network effectively learns residuals. However, it's gain in accuracy comes at a cost of both memory demands as well as execution time. ($\sim102$MB)

\textbf{MobileNet \cite{MobileNets.2017}:} Utilizing the idea of separable convolutions MobileNets manage to offer reduced computational cost with slight degradation in classification accuracy. It applies a single filter at each input channel and then linearly combines them. Thus is designed can be easily parametrized and optimized for mobile applications.

\subsection{Image classification for emergency response and disaster management}\label{subsec:rw}

In this section some relevant works for the problem of aerial image classification for emergency response and disaster management are described, some of which also target remote sensing with UAVs. 

In \cite{Kim.Fire.2016} the authors propose a cloud based deep learning approach for fire detection with UAVs. The detection using a custom convolutional neural network (similar in strcuture to VGG16) which is trained to discriminate between fire and non-fire images of $128\times128$ resolution. The system works by transmitting the video footage from a UAV to a workstation with an NVIDIA Titan Xp GPU where the algorithm is executed. of course, in scenarios with limited connectivity missions there would be difficulties in applying this approach. Overall, the proposed approach achieves an accuracy in the range of $81-88\%$ for this task.

In \cite{Bejiga.Avalanche.2017} a method is proposed for detecting objects of interest in avalanche debris using the pretrained inception Network \cite{SzegedyInception2014} for feature extraction and a linear Support Vector Machine for the classification. They also propose an image segmentation method as a preprocessing technique that is based on the fact that the object of interest is of a different color than the background in order to separate the image into regions using a sliding window. In addition, they apply post-processing to improve the decision of a classifier based on hidden Markov models. The application is executed on a desktop computer and not on an embedded device, with clock speed of 3GHz and 8GB RAM average a performance of 5.4 frames per second for $224\times 224$ images. The accuracy was between $72-97\%$.

Similarly, the work in \cite{Sharma.Fire.2017} also targets fire detection with deep learning. Specifically, two pretrained convolutional neural networks are used and compared, namely VGG16 \cite{VGG.2014} and Resnet50 \cite{ResNet.2015} as base architectures to train fire detection systems. The architectures are adapted by adding fully connected layers after the feature extraction to measure the classification accuracy. The different models average an accuracy of $\sim91\%$ for a custom database with an average processing time of $1.35$ seconds on an NVIDIA GeForce GTX $820$ GPU.

The work in \cite{Zhao.SaliencyFire.2018} proposes an approach comprised of a convolutional neural network called \textit{Fire\_Net} consisting of $15$ layers with an architecture similar to the \textit{VGG16} network with $8$ convolutional, $4$ max-pooling, and $2$ fully connected layers for recognizing fire in $128\times 128$ resolution images. It is accompanied by a region proposal algorithm that extracts image regions from larger resolution images so that they can be classified by the neural network. The training of the system was performed on an NVIDIA GeForce $840$M GPU, while the overall accuracy is $\sim98\%$ and the average performance is $24$ frames-per-second on the particular GPU platform for $128\times 128$ resolution images and without considering the overhead of the region proposal and region selection algorithms.  
In \cite{Kamilaris.DisManUAV.2017} a deep convolutional neural network is trained to classify aerial photos in one of $5$ classes corresponding to natural disasters. The VGG \cite{VGG.2014} network is used as the base feature extraction and a fully connected is placed on top of it to perform the transfer learning for the new task. An accuracy of $91\%$ is achieved for a custom test set and on average less than $3$ seconds are needed to process an image of $224\times224$ on an Intel Core i7 machine.

From the literature analysis it is clear that existing approaches use existing pre-trained networks which adapt through transfer learning for the classification of a single event and primarily utilize desktop-class systems as the main computational platform that remotely process the UAV footage on GPUs. However, in certain scenarios the communication latency and connectivity issues may hinder the performance of such systems necessitating higher autonomy levels for the UAV and on-board processing capabilities \cite{Sze.MLchallOppo.2016}. Also, the computing limitations of embedded platforms constitute the use of existing algorithms targeting desktop-class systems infeasible.

\section{Deep Learning for Aerial Disaster-Event Classification}\label{sec:approach}

This section outlines the process of developing an efficient convolutional neural network suitable for embedded platforms for classifying aerial images from a UAV for emergency response and disaster management applications.

\subsection{Dataset Collection}\label{subsec:aider}

\begin{figure*}[t]
	\centering
	\includegraphics[width=0.9\linewidth]{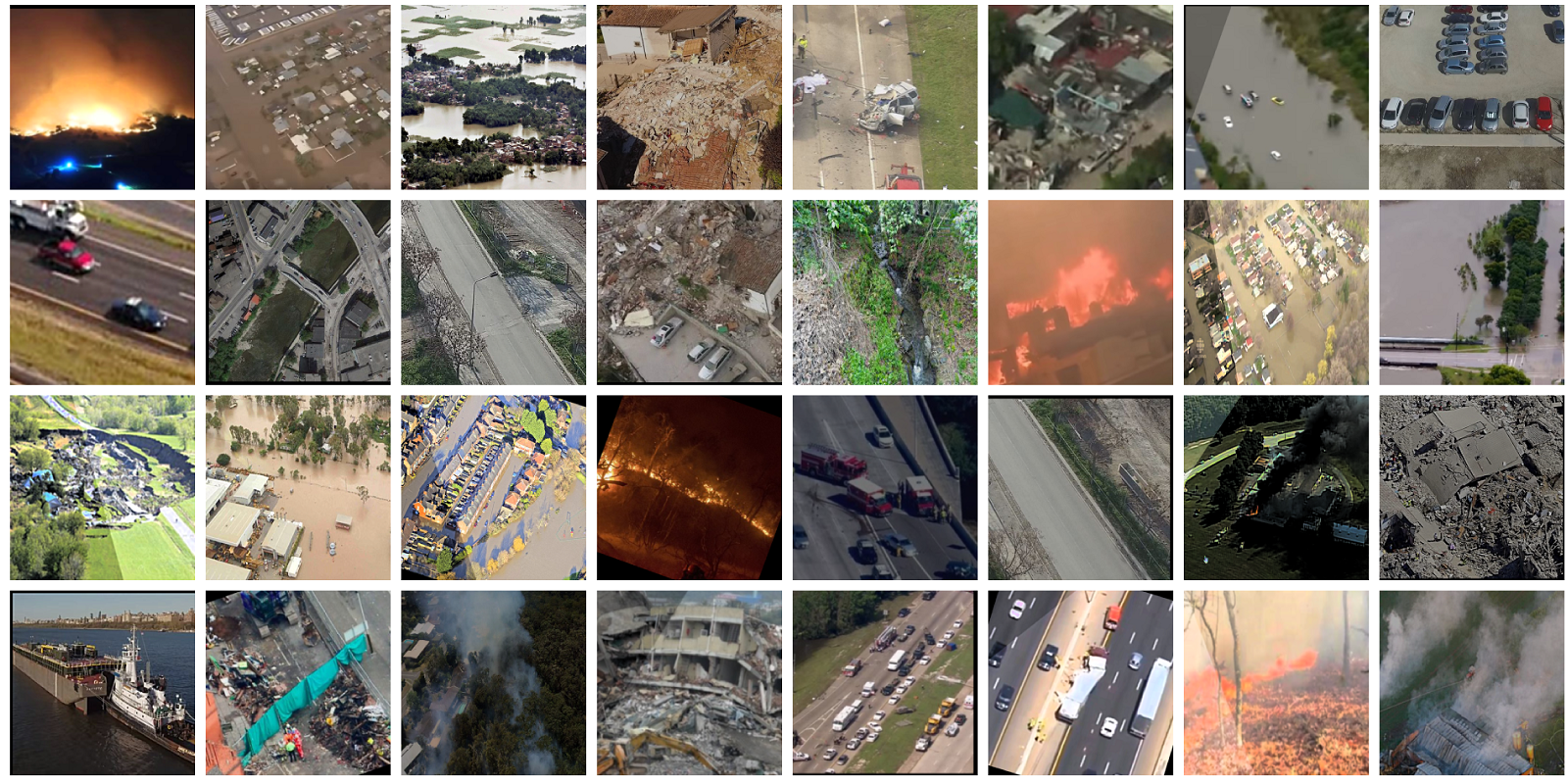}
	\caption{\textbf{A}erial \textbf{I}mage \textbf{D}ataset for \textbf{E}mergency \textbf{R}esponse (AIDER) Applications: Example images from the of Augmented Database}
	\label{fig:dataset}
\end{figure*}

Training a CNN for aerial image classification for emergency response and disaster management applications first requires collecting a suitable dataset for this task. To the best of our knowledge there is no widely used and publicly available dataset for emergency response applications. As such, a dedicated database for this task is constructed referred to as \textit{AIDER} (\textbf{A}erial \textbf{I}mage \textbf{D}ataset for \textbf{E}mergency \textbf{R}esponse Applications). The dataset construction involved manually collecting all images for four disaster events, 320 images of \textit{Fire/Smoke}, 370 images for \textit{Flood}, 320 images for \textit{Collapsed Building/Rubble}, and 335 images for \textit{Traffic Accidents}, as well as 1200 images for the \textit{Normal} case.  Visually similar images such as for example active flames and smoke are grouped together.

The aerial images of these disaster classes were collected from multiple sources such as the world-wide-web (e.g. google images, bing images, youtube, news agencies web sites, etc.), other databases of general aerial images, and images collected using our own UAV platform. During the data collection process the various disaster events were captured with different resolutions and under various condition with regards to illumination and viewpoint. Finally, to replicate real world scenarios the dataset is imbalanced in the sense that it contains more images from the \textit{Normal} class. Of course, this can make the training more challenging, however, a certain strategy is followed to combat this during training which will be detailed in the following sections. 
 
The operational conditions of the UAV may vary depending on the environment, as such it is important that the dataset does not contain only "clean" and "clear" images. In addition, data-collection can be time-consuming and expensive. Hence to further enhance the dataset a number random augmentations are probabilistically applied to each image prior to adding it to the batch for training. Specifically these are geometric transformations such as rotations, translations, horizontal axis mirroring, cropping and zooming, as well as image manipulations such as illumination changes, color shifting, blurring, sharpening, and shadowing. Each transformation is applied with a random probability which is set in such as way to ensure that not all images in a training batch are transformed so that the network does not capture the augmentation properties as a characteristic of the dataset. The objective of all these transformations is to combat overfitting and increase the variability in the training size to achieve a higher generalization capability. Some samples from the dataset can be seen in Fig. \ref{fig:dataset}. Overall, with respect to the related works that consider multiclass problems (e.g., \cite{Kamilaris.DisManUAV.2017}) almost $5\times$ more data were collected. In addition, using augmentations the initial dataset was considerably expanded even further.

\subsection{CNNs for Aerial Disaster Classification}\label{subsec:cnn_models}

To identify the best structure of the CNN that will perform the aerial image classification a number of different networks was developed using two different approaches. The overall objective of this process is to explore the performance-accuracy trade-offs between these networks. First, transfer learning is employed to train the networks outlined in Section \ref{subsec:CNNs} which correspond to the methodology used in prior works. Furthermore, new network structures are designed and trained from scratch specifically for this task. The reasoning behind the latter approach is that it allows making those design choices that lead to more efficient networks that are fast to execute on embedded platforms and at the same time maintain the accuracy of larger networks.

\subsubsection{Transfer Learning Networks}\label{subsubsec:pretrained}
For transfer learning established networks are used, namely \textit{VGG16} \cite{VGG.2014} and \textit{ResNet50} \cite{ResNet.2015} which have also been used in prior works outlined in Section \ref{subsec:rw} such as \cite{Kamilaris.DisManUAV.2017,Sharma.Fire.2017,Kim.Fire.2016} as well as networks more suited to embedded domains such as \textit{MobileNet}\cite{MobileNets.2017}. The feature extraction part is frozen for each of these networks, applying all necessary preprocessing steps to the input image, and add a classification layer on top similar to prior works. In contrast to other works a global (per feature-map) average-pooling layer is applied prior to the dense layers followed by a softmax classification layer at the end. The average pooling reduces the parameter count and the subsequent computational and memory requirements and has shown to perform equally as well with the traditional approaches \cite{Lin:2013:NiN:GAP}. Hence, the pretrained models used for comparison are inherently more efficient in terms of memory and operations that the networks used in the literature for this task, which utilize fully connected layers.

\subsubsection{Custom Networks}\label{subsubsec:custom}

The larger and deeper networks attained through transfer learning may not be suited for resource-constrained systems such as UAV platforms, which impose limitations of the size of the platform, the weight, and its power envelope. For this reason there is a need to design specialized networks that are inherently computationally efficient. The design space is explored by focusing on the layer configurations, type and connectivity. Consequently different networks are developed to better understand the trade-offs involved in the design choices. There are some systematic design choices that are made across the different network configurations. 

\begin{figure*}[t]
	\centering
	\includegraphics[width=0.95\linewidth]{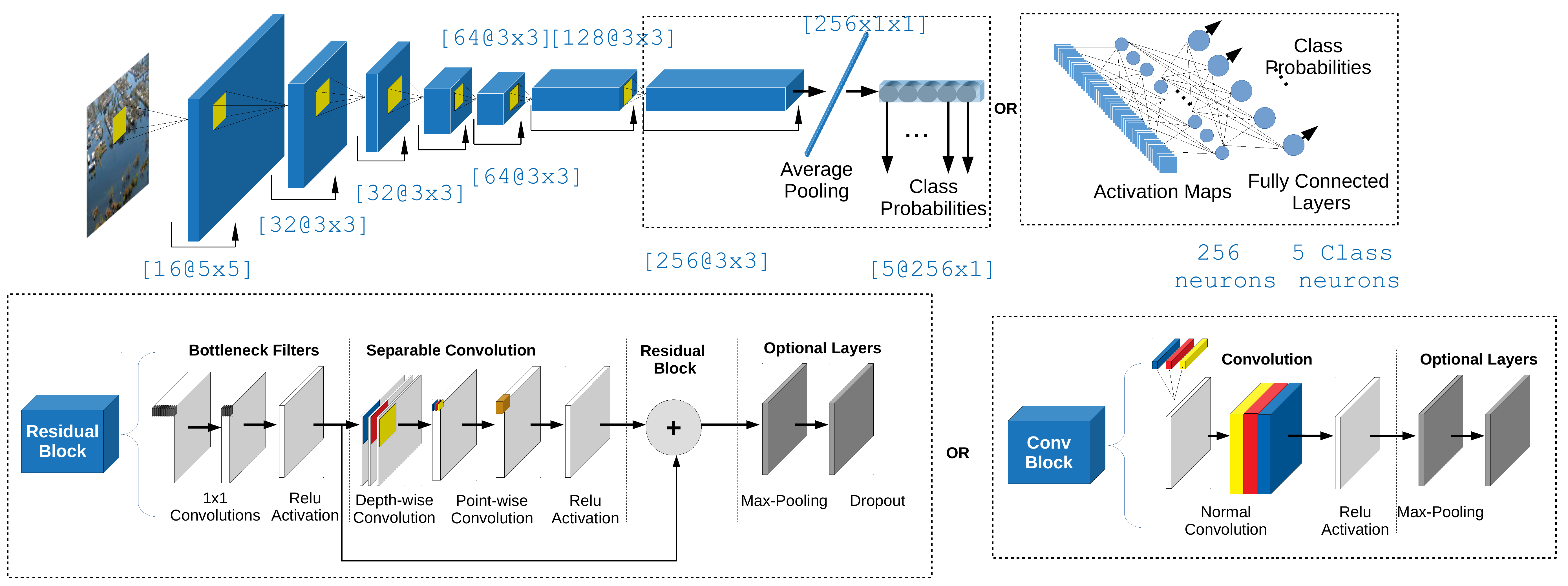}
	\caption{Different configurations for setting up a CNN model for aerial disaster classification. }
	\label{fig:cnn_arch}
\end{figure*}

\begin{itemize}
    \item \textbf{Reduced Cost of First Layer}: The first layer typically incurs the higher computational cost since it is applied on the whole image. Hence, a relatively small number of filters is selected ($16$) with higher spatial resolution of $5\times 5$ compared to the latter layers. Overall, using an increased filter size resulted in an increase in accuracy. The improvement is attributed to the fact that since a relatively small number of filters is used compared to other works (e.g., $64$) more image information by needs to be captured by increasing the filter size. Experiments were  also performed starting with a filter depth of 8 but this resulted in reduced accuracy.
    \item \textbf{Early downsampling}: Max pooling layers are used after each convolution layers (either separable or normal) to half the input size and effectively reduce the computational cost, It was empirically found that downsizing the feature maps in the latter stages resulted in decreased accuracy hence, the downsampling is performed in all but the two final convolutional layers.
    \item \textbf{Canonical Architecture:} To keep the representational expressiveness a pyramid-shaped form is adopted for the CNN configuration, which means a progressive reduction of spatial resolution of the feature maps at each layer with an increase of their depth. It is quite typical for large networks to have even thousands of filters at each layer, however, for embedded applications this adds considerable overhead. Hence, the first layer has $16$ filters, which are then doubled thereafter but do not increase it beyond $256$ which is the final layer prior to the classification part. 
    \item \textbf{Fully Convolutional Architecture:} A simple and effective trick is utilized to massively reduce the parameter count and computational cost by replacing the fully connected layers with a global average pooling operation followed by convolutional layers (before feeding into the softmax activation) for reduced number of parameters.
    \item \textbf{Efficient Regularization:} Due to the relatively small size of the dataset compared to databases such as ImageNet; additional regularization techniques are also incorporated beyond augmentation to combat overfitting. In particular, \textit{batch normalization} [3] is used after each convolutional layer and also the dropout [12] strategy is employed with a ratio of $0.5$ of drop connections during training. All the convolutional layers are also regularized using an L2 regularizer.
    \item \textbf{Network Depth:} Deep networks are necessary to build strong representations but are also predicated on having a huge amount of data. Also very deep networks incur a higher computational cost. Given these two factors it was found that a network size of $7$ major processing blocks (combined layers) was sufficient to achieve comparable accuracy to the state-of-the-art, while increasing it did not result in significant accuracy improvements but incurred higher computational cost.
    \item \textbf{Residual Connections:} Residual learning is incorporated in the architecture. Specifically, skip connections are introduced from the input of a computation block to its output where it is merged back with an element-wise addition. As it will be shown in the experiments residual learning enables to increase the accuracy further with a slight decrease of the computation performance.
\end{itemize}

The aforementioned configurations are combined to build the \textit{ERNet} architecture shown in Fig. \ref{fig:cnn_arch}. In addition, using the aforementioned basic principles different networks are designed featuring various combinations of the main configurations in order to compare and contrast the trade-offs. Specifically, the main differences between the networks stem from the use of specific layer types and techniques such as separable convolutions, and residual connections. First, a canonical CNN (referred to as \textit{baseNet}) is designed composed of normal convolutional layers with fully connected layers at the end. In the second network configuration the convolutional part is replaced by a separable convolution (depth- and point- wise), while maintaining the fully-connected layer at the end (referred to as \textit{SCNet}). For the third network configuration the fully-connected portion is replaced by convolutional layers in order to reduce the parameter size and memory demands of the network even further (referred to as \textit{SCFCNet}). For the final configuration, residual connections are added to each layer to improve the learning performance and this constitutes the final architecture which is referred to as \textit{ERNet}. The progressive changes in the network configurations allows us to study how each choice impacts both performance and accuracy the results of which will be shown in Section \ref{sec:results}.

\subsection{Training}\label{subsec:training}
All the networks are developed and tested through the same framework so as to have the same conditions and a fair comparison during the inference phase. The Keras deep learning framework \cite{keras} is used which has available all the pretrained models used for transfer learning, with Tensorflow \cite{tensorflow} running as the backend \footnote{We plan on releasing all the models and the training data as open source}. The same image size is used for all networks where possible (except for the \textit{mobileNet} which specifically requires a smaller image size). Consequently, before augmenting and adding an image to the batch it is first resized to the appropriate image size depending on the network (default is $240\times240$ pixels which is a typical size for training CNNs). It should be noted that it is possible to use larger image sizes at a cost of slower inference time, however in this work the image size space is not explored but rather focus is on the network design. 

The first step in the training process is to split the dataset into training, validation, and test sets. The bulk of the data are allocated to the training set and the rest between the other two sets in a 0.6, 0.2, 0.2 ratio. As mentioned prior, the \textit{Normal} class is the majority class and thus is over-represented in the dataset. This reflects real-world conditions, however, if not addressed, it can potentially lead to problems where the network overfits and thus classifies everything as the majority class. To avoid issues due to the dataset imbalance the simultaneous use of majority class undersampling with oversampling of the minority classes within the same batch is performed. To do this we select the same number of images form each class to form a batch and this way all cases are equally represented.

All the networks where trained using a GeForce Titan Xp, on a PC with an Intel $i7-7700K$ processor, and $32$GB of RAM. The Adam optimization method was used for training with learning rate-decay starting from a learning rate of $0.001$, and multiplying it by a factor of $0.95$ every $5$ epochs to achieve a smoother decrease. Each network is trained for $200$ epochs each comprising of $100$ batch iterations, with a batch size of $64$ resulting in $6400$ generated training images per epoch.

\section{Experimental Evaluation and Results}\label{sec:results}

In this section the analysis of the trained networks is presented with results from the experimental evaluation of the approach on an actual embedded platform attached on the UAV as well as the UAV ground station. Evaluation is performed on a desktop i7 CPU that can be easily ported to a computational platform used in UAVs such as an Android platform that acts as the mobile control station or embedded devices such as Odroid XU4.

\subsection{Performance Metrics}\label{subsec:metrics}
The ultimate objective of this work is to be able to run the models on board a UAV and process each image online. Hence, an important performance metric is the achievable frame-rate or frames-per-second (FPS) achieved by each model, which is inversely proportional to the time needed to process a single image frame from a video. In addition, since the prior distribution over classes is signiﬁcantly nonuniform a simple accuracy measure (percentage of correctly classiﬁed examples) which is used in related works, may not be appropriate for the specific problem considered in this work since usually the normal case would have a larger number of samples in the test and training set than the other classes. To avoid this bias in our results an average accuracy ($\overline{A}$) metric \cite{Evaluation:Metrics:Classification:2015} is employed instead that averages across the accuracies of each class rather than that of the test set as a whole. 

\begin{table*}[t]\centering
\begin{threeparttable}
  
  \caption{Summary of Results for all the trained models on PC Setup}
  \label{table:model_res}
  
  \begin{tabular}{|c|c|c|c|c|c|c|}
    \hline
    CNN Model & Type$^1$ & Average Accuracy (\%) & Processing Time (ms) $^2$ & Frames-Rate$^2$ & Speedup$^3$ & Memory (MB)\\ \hline
  	\textit{ERNet} & C
  	 & 90.1 & 18.7 & 53 & 18.5 & 0.3 \\ \hline
  	\textit{SCFCNet}  & C
  	 & 87.7 & 13.1 & 76 & 26.4 & 0.2\\ \hline
  	\textit{SCNet} & C
  	& 85.4 & 14.1 & 70 & 24.5 & 6.5\\ \hline
  	\textit{baseNet} & C
  	& 88 & 21.2 & 47 & 16.3 & 7\\ \hline
  	\textit{VGG16} & T
  	& 91.9 & 346 & 2 & 1 & 59.3\\ \hline
  	\textit{ResNet50} & T
  	& 90.2 & 257 & 3 & 1.3 & 96.4\\ \hline
  	\textit{MobileNet} & T
  	& 88.5 & 48.2 & 20 & 7.1 & 13.9\\ \hline
  \end{tabular}
  
  \begin{tablenotes}
    \footnotesize
      \item $^1$ C: Custom Network Trained from scratch | T: Pretrained network used for transfer learning to the new task
      \item $^2$ Processing speed and Frame-Rate as measured on an Intel i7 CPU.
      \item $^3$ Speedup with respect to the network with the higher accuracy which is the \textit{VGG16}.
  \end{tablenotes}

\end{threeparttable}
\end{table*}

\subsection{Overall Performance, Analysis and Comparison}\label{subsec:analysis}
The results for all networks are summarized in Table \ref{table:model_res}. First, with regards to the accuracy of the pretrained models it is observed that \textit{VGG16} outperforms all of them with a $91.9\%$ $\overline{A}$. This is in line with what has been reported in prior works using this network achieving an accuracy between $81-98\%$ for different applications and scenarios however. With regards to the frame-rate it achieves $2$ FPS which is not suitable for real-time use. The fastest of the pretrained networks is \textit{MobileNet} achieving a frame-rate of $20$, however it operates on smaller image resolution ($224\times224$) and achieves a average accuracy of $88.5\%$. Also, all the networks require over $10MB$ which may be prohibitive for on-chip storage in embedded platforms. It is clear from this analysis that it is necessary to investigate tailored made solutions for constrained applications in order to provide an improvement on all design parameters.

Appropriately then the evaluation of the custom networks is presented next. Even starting from a \textit{baseNet} network and following the design choices outlined in Section \ref{subsubsec:custom} it is noticeable that it performs close enough and in some cases outperforms some pretrained networks with regards to average accuracy. Also as a consequence of the careful design choices it manages to offer a $16\times$ speedup over the most accurate network which is the \textit{VGG16}. Applying additional optimizations such as employing separable convolution filters (\textit{SCNet}) can further improve performance in terms of FPS however, it negatively impacts the accuracy as a $\sim3\%$ drop is observed. Surprisingly, using a fully-convolutional approach (\textit{SCFCNet}) mitigates this factor. This can be attributed to the fact that the spatial representations are preserved within the feature maps and do not collapse such as in the case of using fully dense layers. Still, however, the average accuracy is lower than that of the \textit{baseNet} model. The very high frame-rates achievable by the existing models affords us a much larger margin to explore the design space in an attempt to further improve the accuracy. By introducing the final component of our design, the residual connections to form the final \textit{ERNet} architecture it manages to achieve $90\%$ average accuracy which is very close to the pretrained network approaches that have been used in prior works. Furthermore, it achieves over $50$ FPS on a CPU-platform which makes the network suitable for real-time UAV applications. Also importantly, the final memory requirements for the network are $\sim300KB$ which also makes it suitable for on-chip storage on low-power platforms with limited memory as well as more specialized computing platforms such as FPGAs which can have limited on-chip storage.

\begin{figure}[t]
	\centering
	\includegraphics[width=0.5\linewidth]{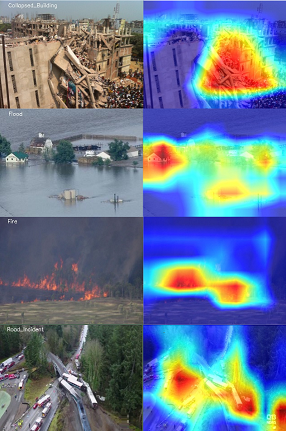}
	\caption{Images classified correctly and the corresponding class activation map. In all cases the visualization shows that the network focuses on important cues within the image to make a decision. From top to bottom: (a) an image of a collapsed building. (b) A flooded area. (c) Forest Fire. (d) Transportation Incident.}
	\label{fig:keras_vis}
\end{figure}

\begin{figure}[t]
	\centering
	\includegraphics[width=0.5\linewidth]{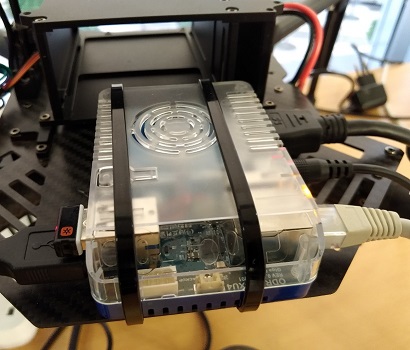}
	\caption{Experimental embedded platform: Odroid-XU4 embedded platform on-board a DJI Matrice 100 UAV}
	\label{fig:embedded_platforms_odroid}
\end{figure}

\subsection{Evaluation of Learning}\label{subsec:learn_eval}
In this section a closer look is taken into what features and image regions influence the prediction of the neural network and what it has learned to respond to in the various cases in order to come to a classification decision. Such an analysis is particularly important to enable transparency as to why the models predict what they do and establish appropriate trust and confidence in users. In addition, such methodologies are also important during the training phase as they enable the identification of failure modes. To this end, we follow the Gradient-weighted Class Activation Mapping (Grad-CAM) approach in \cite{Selvaraju.CAM.2017} to generate heat maps of the image regions that mostly influence a classification decision. Fig. \ref{fig:keras_vis} indicates some images which have been correctly classified and the produced heat-map indicating important regions in the CNN's (the \textit{ERNet} model in particular) decision making. Notice that the CNN uses class-specific features to make a decision. For example, the demolished side of the building gin the first image and the  red-orange glow of fire in the third image. The second and fourth images are more complex to analyze. In the former it seems that the network infers the flood first by the water and then by the presence of buildings that indicate a flooded area instead of a river for example. In the latter image again a combination of cues makes the network come to a decision such as the crashed train and the road segments.

\subsection{Embedded Platform Results}\label{subsec:embedded}
UAV platforms can differ in their implementation depending on the use-case and deployment strategy; from very lightweight with only on-board components to requiring dedicated computing infrastructure at the ground station. The latter case has already been covered through the previous analysis. Thus in this section the focus is on the on-board processing platform. Experimental setup using a DJI Matric $100$ UAV \footnote{https://www.dji.com/matrice100} and experimental results are shown in Fig. \ref{fig:embedded_platforms_odroid}. For the on-board processing the ODROID-XU4 (Fig. \ref{fig:embedded_platforms_odroid}) computing device is chosen which is powerful and energy-efficient and comes in a small form factor suitable for UAVs. \footnote{It features a Samsung Exynos5 Octa ARM Cortex, 2Gbyte LPDDR3 RAM
and uses between 10W and 20W}. The \textit{ERNet} model achieves $\sim9$ FPS on this platform which is already far more practically applicable than other state of the art models that achieve at most $\sim3$ FPS with lower accuracy. In addition, by applying quantization and bit reduction techniques it would be possible to further improve performance for CPU platforms and potentially run at even higher frame-rates.

\section{Conclusions and Future Work}\label{sec:conc}
This paper presented the first steps towards the automated classification of disaster events in real-time from on-board a UAV. It introduced a dedicated aerial image dataset for emergency response applications which researchers can use to further advance the existing models. The dataset will be further expanded and enhanced with additional images and classes in order to further raise the awareness of the community towards such applications and improve on existing models and techniques. Furthermore, a small and efficient convolutional neural network \textit{ERNet} is developed that is up to $3\times$ faster on an embedded platform, requires two orders of magnitude less memory and provides similar accuracy to existing models. Going forward, we are eager to investigate even further architectural design choices such as atrous-convolution layers as well as multi-frame processing to capture temporal information and further imprtove the results. It is also beneficial to study the impact of using different color spaces which may help improve the accuracy even further. Finally, the potential to combine \textit{ERNet} with algorithms that detect people and vehicles as well as additional modalities (e.g., infrared camera) can lead to even more enhanced situational awareness that can provide valuable tool for emergency response and disaster management applications.

\section*{Acknowledgements}
Christos Kyrkou would like to acknowledge the support of NVIDIA Corporation with the donation of the Titan Xp GPU used for this research.

{\small
\bibliographystyle{ieee}
\bibliography{arxiv}
}

\end{document}